\documentclass[letterpaper]{article} 
\usepackage{aaai2026}  

\usepackage{times}  
\usepackage{helvet}  
\usepackage{courier}  
\usepackage[hyphens]{url}  
\usepackage{graphicx} 
\usepackage{natbib}  
\usepackage{caption} 
\usepackage[colorlinks=true, linkcolor=blue, citecolor=blue, urlcolor=blue]{hyperref}

\usepackage{algorithm}
\usepackage{algorithmic}

\usepackage{newfloat}
\usepackage{listings}
\DeclareCaptionStyle{ruled}{labelfont=normalfont,labelsep=colon,strut=off} 
\floatstyle{ruled}
\newfloat{listing}{tb}{lst}{}
\floatname{listing}{Listing}

\usepackage{graphicx}
\usepackage{amsmath}
\usepackage{multirow}
\usepackage{tcolorbox}
\usepackage{xcolor}
\usepackage{enumitem}
\tcbuselibrary{listings, breakable, skins}
\usepackage{longtable}
\usepackage{booktabs}
\usepackage{array}
\usepackage{pifont}
\newtcolorbox{reasoningbox}[2][]{colback=white!98!gray!2, colframe=black!75!gray,
fonttitle=\bfseries, title=#2, #1, breakable, enhanced}

\nocopyright

\title{Graph of Verification: Structured Verification of LLM Reasoning \\ with Directed Acyclic Graphs}

\author {
    Jiwei Fang\textsuperscript{\rm 1},
    Bin Zhang\textsuperscript{\rm 2},
    Changwei Wang\textsuperscript{\rm 3},
    Jin Wan\textsuperscript{\rm 3},
    Zhiwei Xu\textsuperscript{\rm 1}\thanks{Corresponding author: \texttt{zhiwei\_xu@sdu.edu.cn}}
}
\affiliations {
    \textsuperscript{\rm 1} Shandong University \\
    \textsuperscript{\rm 2} The Key Laboratory of Cognition and Decision Intelligence for Complex Systems, Institute of Automation, Chinese Academy of Sciences \\
    \textsuperscript{\rm 3} Qilu University of Technology \\
}

\begin{document}

\maketitle

\begin{abstract}
Verifying the complex and multi-step reasoning of Large Language Models (LLMs) is a critical challenge, as holistic methods often overlook localized flaws. 
Step-by-step validation is a promising alternative, yet existing methods are often rigid. 
They struggle to adapt to diverse reasoning structures, from formal proofs to informal natural language narratives. 
To address this adaptability gap, we propose the Graph of Verification (GoV), a novel framework for adaptable and multi-granular verification. 
GoV's core innovation is its flexible \textit{node block} architecture. 
This mechanism allows GoV to adaptively adjust its verification granularity—from atomic steps for formal tasks to entire paragraphs for natural language—to match the native structure of the reasoning process. 
This flexibility allows GoV to resolve the fundamental trade-off between verification precision and robustness.
Experiments on both well-structured and loosely-structured benchmarks demonstrate GoV's versatility. 
The results show that GoV's adaptive approach significantly outperforms both holistic baselines and other state-of-the-art decomposition-based methods, establishing a new standard for training-free reasoning verification.

\end{abstract}

\begin{links}
    \link{Code}{https://github.com/Frevor/Graph-of-Verification}
\end{links}

\section{Introduction}

Despite the remarkable progress of Large Language Models (LLMs) in complex reasoning tasks~\cite{Achiam2023Gpt, Shao2024Deepseekmath, Yang2024Qwen25TR, grattafiori2024llama}, verifying the fidelity of their reasoning processes remains a fundamental challenge, especially in domains involving intricate mathematics and logic. 
Beyond occasional computational or logical mistakes that lead to incorrect outcomes, a more concerning issue is that LLMs frequently produce reasoning steps that appear plausible yet contain subtle logical flaws, even when the final answer happens to be correct~\cite{He2025CanLL, Luo2024Improve}.
Consequently, there is a critical need for methods that can rigorously and scalably assess the internal validity of LLM-generated reasoning, ensuring the dual correctness of both the process and the outcome.

\begin{figure}[t]
\centering
\includegraphics[width= 0.9\linewidth]{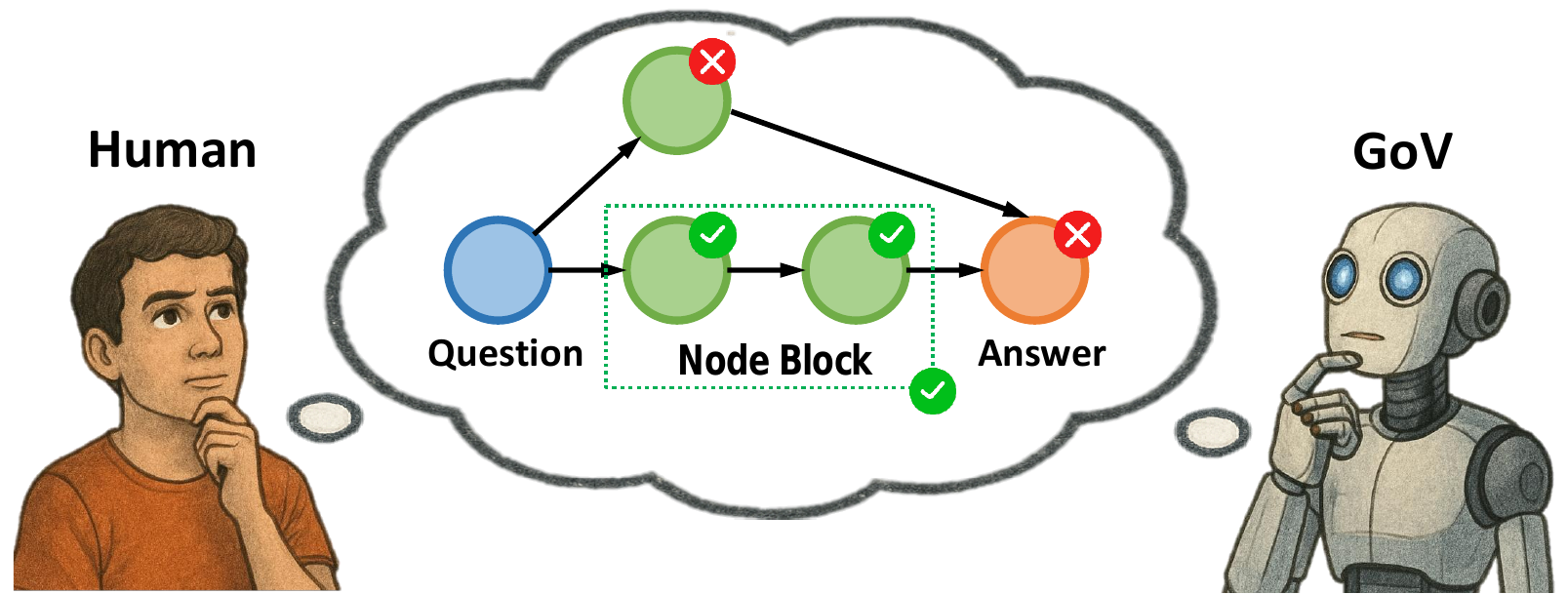}
\caption{Both humans and GoV approach reasoning validation by decomposing the process into a directed acyclic graph. The granularity of these validation units is highly flexible, allowing for adaptation to different tasks.\looseness=-1}
\label{fig:intro}
\end{figure}

To tackle this verification challenge, we draw inspiration from a fundamental aspect of human cognition. 
When faced with a complex argument, humans instinctively decompose it into a series of smaller, dependent steps, verifying the logic sequentially. 
This process can be naturally modeled as a directed acyclic graph (DAG)~\cite{Wavering1989Logical}, as illustrated in Figure~\ref{fig:intro}. 
This intuitive "divide and conquer" strategy stands in stark contrast to the prevailing approach in LLM verification: \textbf{holistic verification}, where the entire reasoning process is evaluated in a single pass. 
Such a monolithic approach often overwhelms the verifier LLM with excessive cognitive load, hindering its ability to detect localized flaws, a vulnerability that becomes particularly pronounced as the reasoning chain lengthens and complexity increases.
We argue that embracing a structured, decomposition-based approach is the indispensable path toward reliable reasoning evaluation.

However, this divide-and-conquer principle reveals a deeper, more fundamental challenge: how should one effectively decompose the ubiquitous, loosely-structured reasoning expressed in natural language? 
While formal proofs can be straightforwardly decomposed along its explicit logical structure, the field currently lacks a unified framework capable of navigating the entire spectrum of reasoning—from the highly structured to the highly informal. 
This adaptability gap is a critical barrier to achieving truly reliable reasoning evaluation.

To address this challenge, we propose \textbf{G}raph \textbf{o}f \textbf{V}erifi-cation (\textbf{GoV}), a novel framework designed specifically for \textbf{adaptability} and \textbf{multi-granular verification}. 
The core innovation of GoV is its flexible \textbf{node block} architecture. 
This mechanism allows the verification granularity to be configured according to the native structure of the reasoning process. 
For well-structured tasks, a node block can be a single atomic statement, enabling maximum precision. 
In contrast, for loosely-structured natural language narratives, it can be a semantically coherent paragraph, ensuring robustness. 
By dynamically selecting the most appropriate granularity, GoV resolves the tension between precision and robustness, offering a unified and effective verification solution for diverse reasoning tasks.\looseness=-1

Our primary contributions are as follows:
\begin{itemize}
    \item To the best of our knowledge, we are the first to formalize decomposed verification as a \textbf{two-dimensional design space} characterized by \emph{Verification Granularity} and \emph{Contextual Scope}, offering a principled perspective for verifying both well-structured and loosely-structured text.\looseness=-1
    \item We propose \textbf{GoV}, a novel and flexible multi-granular verification framework. Its core node block mechanism enables the verification strategy to be adapted to the structural characteristics of the task.
    \item We demonstrate the effectiveness and generality of GoV through experiments on both well-structured and loosely-structured tasks, showcasing its powerful ability to solve diverse verification challenges in different configurations.
\end{itemize}


\section{Related Work}
\subsection{Reasoning in Large Language Models}
Complex reasoning is a hallmark of advanced LLMs, a capability primarily validated on mathematical and scientific reasoning benchmarks that have become the principal litmus test for their power. 
A key advancement in this area is Chain-of-Thought prompting~\cite{wei2022chain}, which elicits a step-by-step reasoning process from the model, significantly improving its performance on tasks requiring complex planning and computation. 
Building on this, subsequent research has explored more sophisticated reasoning structures, such as exploring multiple reasoning paths with Tree-of-Thoughts~\cite{Yao2023TreeOT} or modeling non-linear dependencies with Graph-of-Thoughts~\cite{Besta2023GraphOT}. 
This body of work has established a clear consensus: a structured reasoning process is crucial for enhancing the quality of solutions generated by LLMs. 
Our work builds upon this principle, shifting the focus from \textbf{generating} structured reasoning to \textbf{validating} it.

\subsection{Verification of LLM Reasoning}
Although generating more elaborate reasoning processes is beneficial, systematically verifying their correctness remains a significant challenge~\cite{Hong2023ACL,Stechly2024OnTS,Gu2024ASO,Zheng2023JudgingLW}. 
To address this issue, existing verification approaches can be broadly categorized as follows.

\paragraph{Holistic Verification and Training-Based Verifiers} 
A common approach is holistic verification, where an LLM is prompted to evaluate an entire reasoning chain at once. 
However, this method often falters when dealing with long and complex reasoning chains, as it imposes a heavy cognitive load on the verification model. 
Another paradigm involves training-based verifiers, most notably process-based reward models~\cite{Lightman2023LetsVS, Wang2023MathShepherdVA, Zhang2025TheLO}, which are fine-tuned on human-annotated data to score or classify intermediate reasoning steps. 
While effective, these models require substantial labeled data, are costly to train, and may struggle to keep pace with the rapidly evolving capabilities of frontier LLMs, necessitating expensive retraining cycles. 
In contrast, our Graph-of-Verification (GoV) framework is \textbf{training-free}, making it a more agile and resource-efficient solution.

\paragraph{Decomposition-Based Verifiers} To overcome the limitations of holistic evaluation, an emerging and powerful trend is decomposition-based verification, which breaks down the reasoning chain into smaller, more manageable units for stepwise validation. 
Recent works, such as PARC~\cite{mukherjee2025premiseaugmented} and Deductive Verification~\cite{Ling2023DeductiveVO}, have demonstrated the potential of this approach. 
These methods typically aim for highly granular verification, focusing on validating atomic reasoning steps against the minimal necessary premises. 
This is achieved either through post-hoc premise extraction or by constraining the generation process to a "natural program" format that explicitly declares premises.

However, the pursuit of atomic granularity and minimal premises, while theoretically ideal, relies on fragile premise extraction or declaration steps that are often unreliable in loosely structured natural language contexts. 
In contrast, GoV is the first framework explicitly designed to address and navigate this challenge. 
By introducing a multi-granularity architecture with node blocks, GoV can adapt its strategy—from fine-grained verification for well-structured tasks to a more robust, coarse-grained approach for natural language—thereby offering a more unified and practical solution to reasoning verification.

\begin{figure*}[t]
\centering
\includegraphics[width= \textwidth]{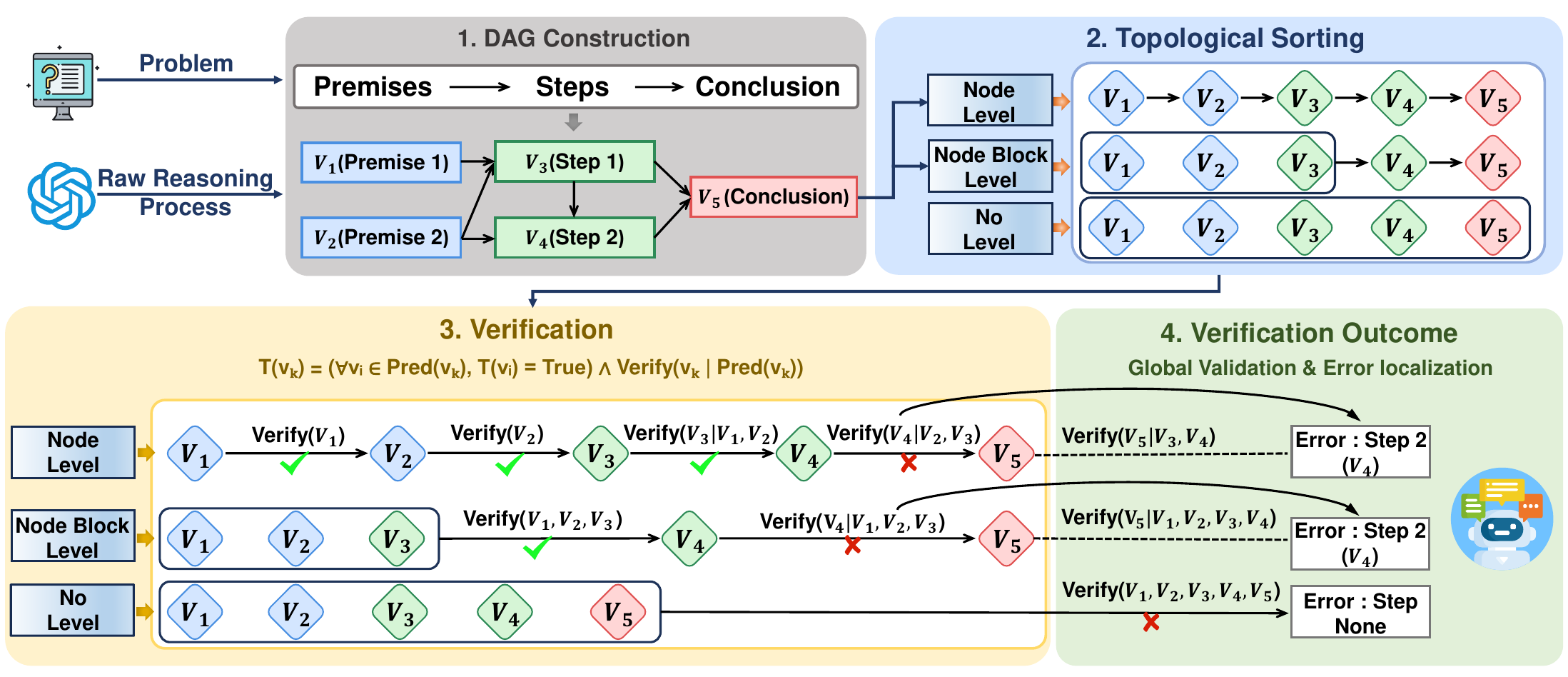}
\caption{The GoV Four-Stage Verification Pipeline. GoV operationalizes structured validation by first modeling reasoning as a (1) Directed Acyclic Graph (DAG). A (2) topological sort then enforces causal consistency, ensuring premises are verified before conclusions. Following this order, (3) sequential verification is performed by an LLM which assesses each unit based on previously validated antecedents. The process (4) terminates at the first detected error, enabling precise fault localization and guaranteeing the logical soundness of any fully verified reasoning chain.
\looseness=-1}
\label{fig:gov_framework}
\end{figure*}

\section{Graph of Verification}

In this section, we first establish the conceptual foundation of our work by defining the design space for decomposed verification. 
This framing clarifies the core trade-offs that any such verification strategy must navigate.
We then formally present the Graph of Verification (GoV), a framework whose components and mechanisms are specifically designed to operate flexibly within this space. 
The overall framework of GoV is illustrated in Figure~\ref{fig:gov_framework}.

\subsection{The Design Space of Decomposed Verification}
\label{sec:gov_design}

While the divide-and-conquer principle is central to verifying complex reasoning, the critical question of how to best decompose a reasoning chain remains underexplored. 
We posit that any decomposition strategy can be characterized within a novel two-dimensional design space: Verification Granularity and Contextual Scope.

\paragraph{Dimension 1: Verification Granularity}
The first dimension, verification granularity, defines the scale of the reasoning unit being validated. This dimension spans a spectrum:
\begin{itemize}[noitemsep, topsep=0pt, leftmargin=*]
    \item \textbf{Atomic Granularity:} On the fine-grained end, the reasoning is decomposed into the smallest possible, indivisible logical or computational units (e.g., a single arithmetic operation '\texttt{a + b = c}'). This approach offers the highest precision and makes error localization exact, but it imposes strict requirements on the structural clarity of the reasoning process.
    \item \textbf{Block Granularity:} On the coarse-grained end, multiple related atomic steps are aggregated into a single, semantically coherent unit for verification (e.g., a full paragraph explaining a particular stage of a solution). This approach is more robust and adaptable to the native structure of natural language reasoning.
\end{itemize}

\paragraph{Dimension 2: Contextual Scope}
Contextual scope defines how much historical information is provided as the premise for verifying the current unit. This also exists on a spectrum:
\begin{itemize}[topsep=0pt, leftmargin=*]
    \item \textbf{Minimal Context:} This approach provides only the theoretically necessary direct premises required to validate the current unit. Its advantage is minimizing cognitive load and reducing distraction, but it relies on a premise extraction process that can itself be fallible.
    \item \textbf{Inclusive Context:} This approach provides a broader set of all previously verified information as context. It is a safer and more robust strategy but risks introducing redundant information.
\end{itemize}
The optimal configuration within this design space is \textbf{NOT} universal; rather, it is contingent on the inherent structure of the reasoning process itself. 
For \emph{well-structured reasoning}, such as formal proofs or programmatic calculations, a strategy of \emph{Atomic Granularity} and \emph{Minimal Context} is often viable and offers the highest precision. 
However, for the more common case of \emph{loosely-structured reasoning} found in natural language narratives, this strategy becomes brittle. 
The difficulty in reliably performing atomic decomposition and premise extraction necessitates a more robust approach, typically favoring \emph{Block Granularity} and \emph{Inclusive Context}. 
The lack of a framework that can flexibly navigate these trade-offs and adapt to this structural spectrum is a critical gap in current research. 
The GoV framework, which we introduce next, is designed precisely to fill this gap.

\subsection{Reasoning with Directed Acyclic Graphs}

The GoV framework is built upon the foundational principle of modeling any reasoning process as a Directed Acyclic Graph (DAG). We adopt this structure because its alignment with the directional and non-circular nature of logical deduction provides the ideal backbone for systematic, decomposed verification.
Reasoning progresses naturally from foundational premises to derived conclusions, a characteristic effectively represented by the directed edges of a DAG. 
This \textit{directionality} ensures that each inference builds upon previously established truths, thereby eliminating ambiguity. 
Moreover, the \textit{acyclic} property of the DAG guarantees well-founded reasoning by removing circular dependencies, thus preventing situations where the validity of a statement relies, whether directly or indirectly, on itself.

Therefore, GoV posits that any reasoning process $\mathcal{S}$ can be formally modeled as a DAG $\mathcal{G}=(V, E)$, where $\mathcal{G}$ serves as the backbone for systematic verification. 
We now provide a detailed description of the two components that constitute the graph.

\paragraph{Node} The set of nodes $V = \{v_1, v_2, \ldots, v_n\}$ in $\mathcal{G}$ comprises the fundamental verification units within the reasoning process. 
Each node $v_i \in V$ typically corresponds to a distinct statement, such as an initial premise, an axiom, an intermediate derived conclusion, or a final assertion. These statements can be broadly categorized:
\begin{itemize}[topsep=0pt, leftmargin=*]
    \item \textbf{Foundational Elements (Premises, Axioms, Facts):} These are the exogenous inputs or fundamental truths forming the starting points of reasoning, often extracted from the problem statement or established domain knowledge.
    \item \textbf{Inferred Statements (Intermediate Conclusions):} These represent new propositions derived from foundational elements or previously validated inferred statements through the application of logical rules, mathematical operations, or other inferential steps.
    \item \textbf{Terminal Statements (Final Conclusions, Goals):} These signify the ultimate outputs or target states the reasoning process aims to establish.\looseness=-1
\end{itemize}

\paragraph{Edge} The set of directed edges $E \subseteq V \times V$ encodes the direct logical dependencies or inferential links between these statements. 
An edge $(v_i, v_j) \in E$ signifies that statement $v_i$ serves as a \textit{direct} premise or justification for the derivation or assertion of statement $v_j$. 
Consequently, the epistemic validity of $v_j$ is contingent upon the validity of $v_i$ (and potentially other co-premises) and the soundness of the inferential step $(v_i, v_j)$ connecting them.

In summary, by modeling reasoning as a DAG, the GoV framework ensures a transparent, structured, and non-circular verification process, where each statement is grounded in a traceable chain of valid inferences.

\subsection{Multi-Granular Topological Units}

To realize the adaptable granularity required to navigate the design space, GoV introduces two primary types of verification units: Atomic Nodes and Node Blocks. 
This multi-granular architecture is a crucial design feature of GoV. 
Before detailing these units, we first establish the verification sequence using a topological sort of the graph: for any DAG $\mathcal{G}$, there exists a vertex ordering $\sigma: V \to \{1, 2, \ldots, n\}$ such that $\sigma(v_i) < \sigma(v_j)$ holds for every edge $(v_i, v_j) \in E$. 
The derived verification sequence $\mathcal{C}_{\text{verif}} = (c_1, c_2, \ldots, c_n)$, where $\sigma(c_k)=k$, ensures each statement is verified strictly after all its logical antecedents, thereby preventing dependency violations.

With the verification order established, we now define the units themselves:

\noindent\textbf{Atomic Nodes:} At the finest granularity, atomic node verification treats each vertex $v_k \in V$ as an independent unit. This mode achieves maximum error localization capability through complete stepwise decomposition.

\noindent\textbf{Node Blocks:} While atomic decomposition offers maximum precision, it is often brittle and impractical for loosely-structured natural language, as argued in our design space framework. GoV's Node Block abstraction directly addresses this by grouping topologically consecutive nodes into logically coherent units $\mathcal{B} = (B_1, \ldots, B_m)$:
\begin{equation}
    \mathcal{C}_{\text{verif}} = (
    \underbrace{c_1,\ldots,c_{|B_1|}}_{B_1},
    \ldots,
    \underbrace{c_{n-|B_m|+1},\ldots,c_n}_{B_m}
    ).
\end{equation}
This coarser granularity reduces verification invocations to $m\leq n$ while enforcing two critical constraints: (i) Topological consistency: for any $B_j, B_K$ with $j<k$, $\max_{v\in B_j} \sigma(v) < \min_{v\in B_k} \sigma(v)$, preserving macro-level dependency relationships; (ii) Semantic cohesion: each block must correspond to complete logical units (e.g., lemmas in proofs or paragraphs in narratives) to prevent arbitrary fragmentation.
Notably, when $m=1$ (i.e., the entire reasoning DAG is treated as a single verification block), the framework degenerates to conventional monolithic verification approaches. 

This adaptable architecture of Atomic Nodes and Node Blocks is the core mechanism that allows GoV to be flexibly configured. 
By choosing the appropriate granularity, users can position their verification strategy at the optimal point within the design space described, balancing precision against robustness to suit the specific reasoning format.

\subsection{GoV Verification Mechanism}
\label{sec:gov_details}

Having defined the multi-granular units of GoV, we now detail the verification mechanism itself. The verification of each unit (atomic node or node block) within the GoV framework is performed by an LLM, guided by two core principles: 
(i) respecting the topological order imposed by $\mathcal{C}_{\text{verif}}$ or its block-wise partitioning $\mathcal{B}$, and (ii) ensuring the availability of all requisite premises that have been previously verified.
The verification function, applied to a target unit, either a node $c_k$ or a block $B_j$, is defined as $\text{Verify}(\cdot,\ Pred_{\text{prov}})$, which determines the validity of the unit based on the provided premises $Pred_{\text{prov}}$, and returns a Boolean outcome (True or False).
The implementation of GoV involves verifying each individual node or node block, and further extends this mechanism to support holistic reasoning across the entire inference process.

\paragraph{Verifying an Atomic Node $c_k$}
The set of \textit{direct logical predecessors} (parents) for an atomic node $c_k$ is $Pred(c_k) = \{c_i \mid (c_i, c_k) \in E\}$.
The premises provided to the LLM, $Pred_{\text{prov}}(c_k)$, must at least include all nodes in $Pred(c_k)$. 
Moreover, every node in $Pred_{\text{prov}}(c_k)$ must have been previously verified as True and must appear before $c_k$ in the topological ordering $\mathcal{C}_\text{verif}$.
That is, $Pred_{\text{prov}}(c_k) \subseteq \{c_1, \ldots, c_{k-1}\}$ where $T(c_i) = \text{True}$ for all $c_i \in Pred_{\text{prov}}(c_k)$.
\begin{itemize}[noitemsep,topsep=0pt,leftmargin=*]
    \item If $Pred(c_k) = \emptyset$ (i.e., $c_k$ is a foundational element), its validity $T(c_k)$ is evaluated directly as $T(c_k) = \text{Verify}(c_k, \emptyset)$, typically by validating it against the problem statement or established knowledge.
    \item If $Pred(c_k) \neq \emptyset$, then $c_k$ is a derived statement. Its validity $T(c_k)$ depends both on the validity of all its immediate predecessors and the soundness of the inference process itself, formally expressed as: 
    \begin{equation}
    \begin{aligned}
        T(c_k) = & \big(\forall c_i \in Pred(c_k), T(c_i) = \text{True}\big) \\
        & \land \text{Verify}\big(c_k, Pred_{\text{prov}}(c_k)\big).
    \end{aligned}
    \end{equation}
\end{itemize}

\paragraph{Verifying a Node Block $B_j$}
For a node block $B_j$, the minimal set of \emph{external} direct prerequisite nodes is defined as:
\begin{equation}
    Pred_{\text{ext}}(B_j) = \Big(\bigcup_{v \in V(B_j)} Pred(v)\Big) \setminus V(B_j).
\end{equation} 
The set of premises provided to the LLM, denoted as $Pred_{\text{prov}}(B_j)$, must cover all information corresponding to $Pred_{\text{ext}}(B_j)$ and consist solely of previously verified blocks that precede $B_j$ in the topological ordering.
In practical scenarios where node blocks correspond to natural language paragraphs (such as in the ProcessBench experiments), $Pred_{\text{prov}}(B_j)$ typically includes the full content of all preceding validated blocks $(B_1, \ldots, B_{j-1})$.
The LLM then performs $\text{Verify}(B_j, Pred_{\text{prov}}(B_j))$ to assess the internal coherence of $B_j$ and validate all statements $c_k$ within $B_j$ based on both the provided premises and any intermediate conclusions derived within the block.

\paragraph{Overall Verification and Error Handling}
The verification process proceeds sequentially along $\mathcal{C}_{\text{verif}}$ (or $\mathcal{B}$). 
If any unit $c_k$ (or block $B_j$) is found to be False, it is identified as the earliest point of failure, and the GoV process typically halts, recording this failure. 
The entire reasoning trajectory $\mathcal{S}$ is considered valid if and only if all its constituent units (nodes or blocks) are successfully verified as True. \looseness=-1

In summary, GoV allows the verification to be conducted on fine-grained atomic nodes or on coarser-grained node blocks (formed from these nodes), depending on the user's definition tailored to the specific task.
In addition, GoV permits a degree of flexibility in selecting $Pred_{\text{prov}}$, allowing the inclusion of a validated superset of direct prerequisites, provided they precede the current unit, when such extended context is expected to facilitate the LLM's assessment. 
This structured, iterative, and premise-driven scheme makes GoV's reasoning validation highly reliable.


\section{Experiments}

To evaluate the practical effectiveness of GoV, we conduct empirical studies on two contrasting reasoning scenarios: one with well-structured, formally defined arithmetic dependencies, and another with loosely-structured, natural language explanations.
These two case studies serve complementary purposes: the first highlights GoV’s capability for fine-grained, high-precision error localization in mathematically rigid tasks, while the second demonstrates its robustness and adaptability in the more challenging setting of free-form reasoning.
Together, they showcase GoV’s versatility and superiority in both deterministic and open-ended verification environments.

To mitigate the influence of model randomness, all models are configured with a temperature of \textit{zero}. 
Furthermore, to comprehensively assess verification performance, we employ the following evaluation metrics:
\begin{itemize}[noitemsep,topsep=0pt,leftmargin=*]
    \item \textbf{Correct Accuracy}\quad The proportion of fully correct reasoning processes that are judged as “correct” by the model.
    \item \textbf{Error Accuracy}\quad The proportion of reasoning processes containing errors in which the model successfully identifies the first erroneous step.
    \item \textbf{F1}\quad The harmonic mean of correct accuracy and error accuracy, serving as the primary metric for evaluating overall model performance.
\end{itemize}

\subsection{Number Triangle Summation}

\begin{figure}[t]
\centering
\includegraphics[width= \linewidth]{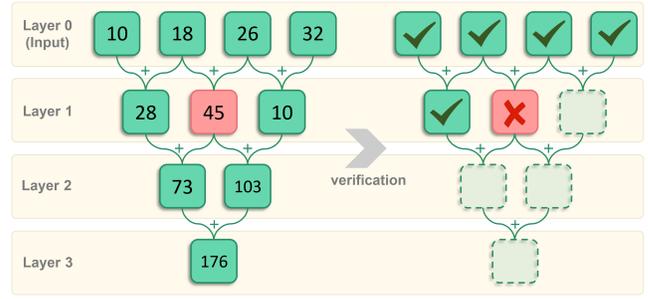}
\caption{Illustration of the numerical validation procedure for Number Triangle Summation.\looseness=-1}
\label{fig:nts_framework}
\end{figure}

To validate GoV's effectiveness in a well-structured reasoning environment, we first introduce the Number Triangle Summation benchmark. 
This task is designed to assess verification accuracy on complex, multi-step reasoning where the dependency graph is explicit and unambiguous.
An illustration of the corresponding verification task is shown in Figure~\ref{fig:nts_framework}.
The reasoning process is defined as follows: starting from $N$ initial numbers at the zeroth layer, each subsequent layer is formed by summing every pair of adjacent numbers from the previous layer. 
This process continues iteratively until a single number remains at the final layer.
Notably, the verification order inherent in this task can be naturally represented as a DAG. 
Verifying the entire reasoning process is inherently long and complex, as the correctness of each intermediate result relies heavily on preceding inference steps.
Existing methods often struggle to reliably verify such multi-step reasoning chains, or fail to accurately localize the specific point at which the reasoning breaks down.

To validate GoV's effectiveness and scalability in a well-structured reasoning environment, we introduce the Number Triangle Summation benchmark. To analyze how verification performance changes with problem complexity, we systematically vary the number of initial inputs ($N$), which determines the length and depth of the reasoning chain. We generated datasets for $N \in \{2, 4, 6, 8\}$. For each value of $N$, we programmatically generated 500 instances. The base layer (Layer 0) of each instance consists of $N$ randomly sampled positive integers between $1000$ and $9999$. The solution steps for each subsequent layer were automatically derived by the program.
To simulate real-world errors, each reasoning process had a 50\% chance of being corrupted by a single-digit perturbation in one of the addition results. This design allows us to directly measure the performance of verification methods as the cognitive load increases.

Given the task's explicit mathematical structure, we configured GoV to operate at the optimal point in our defined design space for such problems: Atomic Granularity for maximum precision and Minimal Context to reduce cognitive load on the verifier LLM. Specifically:
\begin{itemize}
    \item \textbf{Atomic Granularity:} Each individual addition equation is treated as an atomic node in the verification graph. This allows for the most fine-grained error localization possible.\looseness=-1
    \item \textbf{Minimal Context:} During the verification of each node, we provide only its direct prerequisites---the two operands from the previous layer---as the contextual input (${Pred}_{\text{prov}}(v_k)$). This minimizes the cognitive load on the verifier LLM by excluding all irrelevant information.
\end{itemize}

The verification process then proceeds along the natural top-down hierarchy of the triangle, which corresponds to a topological sort of the graph, halting at the first detected error.
We conduct experiments using different LLMs to compare the performance of the proposed GoV framework with that of baseline approaches.

\begin{table}[h]
\centering
\resizebox{\linewidth}{!}{%
\begin{tabular}{llcccc}
\specialrule{1.5pt}{0pt}{0pt}
\multirow{2}{*}{\textbf{Model}} & \multirow{2}{*}{\textbf{Method}} & \multicolumn{4}{c}{\textbf{F1 Score by Problem Size (N)}} \\ 
\cmidrule(lr){3-6}
 &  & \textbf{N=2} & \textbf{N=4} & \textbf{N=6} & \textbf{N=8} \\ \specialrule{1.15pt}{0pt}{0pt}

\multirow{2}{*}{\begin{tabular}[c]{@{}l@{}}Qwen2.5-7B\\-Instruct\end{tabular}} 
& Holistic Verification & 82.3 & 89.5 & 72.8 & 51.5 \\
& GoV (Ours) & \textbf{97.1} & \textbf{89.6} & \textbf{84.8} & \textbf{70.7} \\
\midrule

\multirow{2}{*}{\begin{tabular}[c]{@{}l@{}}Qwen2.5-14B\\-Instruct\end{tabular}} 
& Holistic Verification & 97.1 & 94.2 & 70.8 & 62.9 \\
& GoV (Ours) & \textbf{99.0} & \textbf{95.1} & \textbf{93.9} & \textbf{86.8} \\
\midrule

\multirow{2}{*}{\begin{tabular}[c]{@{}l@{}}Qwen2.5-32B\\-Instruct\end{tabular}} 
& Holistic Verification & \textbf{100.0} & 97.0 & 80.6 & 56.3 \\
& GoV (Ours) & 99.8 & \textbf{99.4} & \textbf{95.8} & \textbf{91.4} \\
\midrule

\multirow{2}{*}{\begin{tabular}[c]{@{}l@{}}Qwen2.5-72B\\-Instruct\end{tabular}} 
& Holistic Verification & \textbf{99.6} & 93.3 & 46.3 & 49.5 \\
& GoV (Ours) & 98.7 & \textbf{97.9} & \textbf{98.1} & \textbf{98.1} \\

\specialrule{1.5pt}{0pt}{0pt}
\end{tabular}%
}
\caption{F1 scores on the Number Triangle Summation task with varying problem sizes ($N$).}
\label{tab:nts_res}
\vspace{-12.5 pt}
\end{table}

\begin{figure}[h]
\centering
\includegraphics[width= \linewidth]{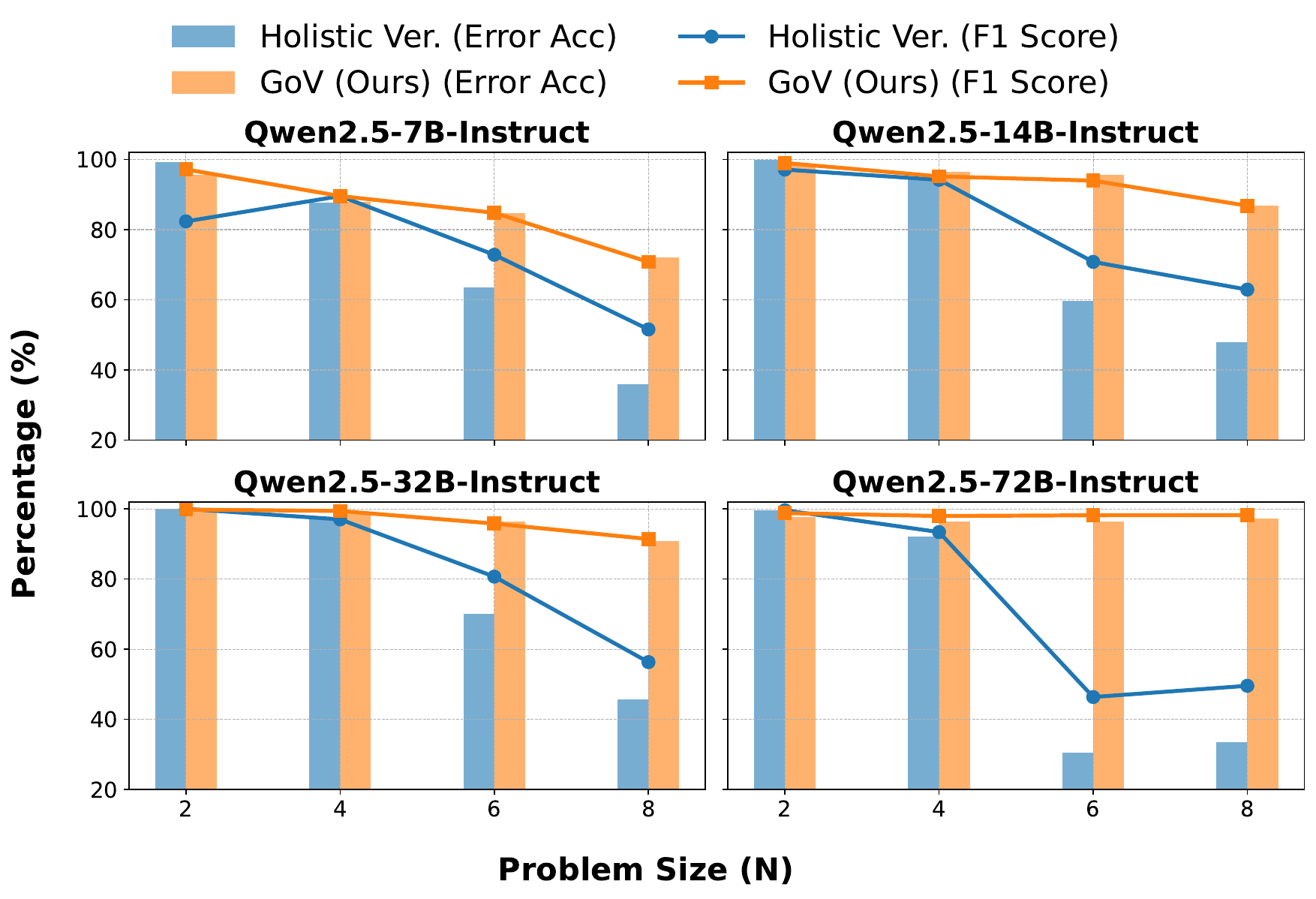}
\caption{The performance of GoV and holistic verification on Number Triangle Summation.\looseness=-1}
\label{fig:F1}
\end{figure}

\begin{table*}[t]
\centering
\resizebox{\linewidth}{!}{
\begin{tabular}{ccccccccccc}
\specialrule{1.5pt}{0pt}{0pt}
\multirow{2}{*}{\textbf{Model}} & \multirow{2}{*}{\textbf{Method}} & \multicolumn{3}{c}{\textbf{GSM8K}} & \multicolumn{3}{c}{\textbf{MATH}} & \multicolumn{3}{c}{\textbf{OlympiadBench}} \\ \cmidrule(lr){3-5} \cmidrule(lr){6-8} \cmidrule(lr){9-11}
 &  & Correct Acc & Wrong Acc & \textbf{F1} & Correct Acc & Wrong Acc & \textbf{F1} & Correct Acc & Wrong Acc & \textbf{F1} \\ \specialrule{1.15pt}{0pt}{0pt}
\multirow{3}{*}{\begin{tabular}[c]{@{}c@{}}Qwen2.5-7B\\ -Instruct\end{tabular}} 
 & Holistic Verification & 66.3 & 36.7 & 47.3 & 63.8 & 23.7 & 34.6 & 46.0 & 25.4 & 32.7 \\
 & PARC & 60.1 & 38.6 & 47.0 & 45.6 & 41.2 & 43.3 & - & - & - \\
 & GoV (Ours) & 89.1 & 43.0 & \textbf{58.0} & 80.7 & 41.7 & \textbf{55.0} & 74.9 & 30.1 & \textbf{42.9} \\ \hline
\multirow{2}{*}{\begin{tabular}[c]{@{}c@{}}Qwen2.5-14B\\ -Instruct\end{tabular}} & Holistic Verification & 93.8 & 47.8 & 63.3 & 86.9 & 40.4 & 55.2 & 76.4 & 30.9 & 44.0 \\
 & GoV (Ours) & 92.2 & 60.3 & \textbf{72.9} & 90.6 & 56.0 & \textbf{69.2} & 83.4 & 42.3 & \textbf{56.2} \\ \hline
\multirow{3}{*}{\begin{tabular}[c]{@{}c@{}}Qwen2.5-32B\\ -Instruct\end{tabular}} & Holistic Verification & 97.9 & 43.0 & 59.8 & 95.6 & 33.3 & 49.4 & 90.0 & 22.4 & 35.9 \\
 & PARC & 95.9 & 55.1 & 70.0 & 86.9 & 53.9 & 66.5 & - & - & - \\
 & GoV (Ours) & 94.3 & 66.6 & \textbf{78.1} & 89.1 & 61.4 & \textbf{72.7} & 79.9 & 55.0 & \textbf{65.2} \\ \hline
\multirow{3}{*}{\begin{tabular}[c]{@{}c@{}}Qwen2.5-72B\\ -Instruct\end{tabular}} & Holistic Verification & 98.4 & 61.4 & 75.6 & 91.9 & 45.3 & 60.7 & 88.5 & 33.7 & 48.9 \\
 & PARC & 97.8 & 59.7 & 74.1 & 86.7 & 53.9 & 66.5 & - & - & - \\
 & GoV (Ours) & 98.9 & 67.6 & \textbf{80.3} & 90.3 & 64.4 & \textbf{75.2} & 84.0 & 49.4 & \textbf{62.2} \\ \hline
\multirow{2}{*}{\begin{tabular}[c]{@{}c@{}}Llama3.3-70B\\ -Instruct\end{tabular}} & Holistic Verification & 96.9 & 66.2 & \textbf{78.6} & 93.1 & 38.4 & 54.4 & 90.0 & 30.9 & 46.0 \\
 & GoV (Ours) & 92.2 & 68.6 & \textbf{78.6} & 79.3 & 65.1 & \textbf{71.5} & 64.6 & 57.0 & \textbf{60.5} \\ \hline
\multirow{2}{*}{GPT4.1-Nano} & Holistic Verification & 89.1 & 44.9 & 59.7 & 75.8 & 55.2 & 63.9 & 71.1 & 47.3 & 56.8 \\
 & GoV (Ours) & 92.2 & 62.3 & \textbf{74.3} & 79.3 & 67.0 & \textbf{72.6} & 67.2 & 57.6 & \textbf{62.0} \\ \specialrule{1.5pt}{0pt}{0pt}
\end{tabular}
}
\caption{Performance comparison on the ProcessBench benchmark. GoV achieves superior F1 scores compared to holistic and PARC baselines consistently across all models and datasets. Results for PARC are sourced from the authors' public rebuttal, as its code was not available at the time of our experiments}
\label{tab:process_res}
\end{table*}

As shown in Table~\ref{tab:nts_res} and Figure~\ref{fig:F1}, all models based on the GoV framework demonstrate substantial improvements in F1 scores. 
Under the baseline setting (CoT-based holistic verification), although the prompt instructs the LLM to examine each equation individually and the generated reasoning chains seem to reflect such step-by-step checking, the verification remains unfaithful.
This observation is consistent with the conclusions reported in the previous literature~\cite{QuestionDecomposition, lindsey2025biology}.

 In contrast, GoV's success can be directly attributed to its configured strategy, which is perfectly matched to the task's structure. By operating at an atomic granularity, it decomposes the complex task into a series of simple, independently verifiable checks. Concurrently, by providing only a minimal context, it eliminates distracting information and allows the verifier LLM to function with maximum reliability. This experiment serves as a powerful validation of our framework's core principle: when a task's structure permits, a high-precision configuration of GoV can achieve outstanding verification performance.

\subsection{ProcessBench}

Having validated GoV's high-precision capabilities in a well-structured environment, we now evaluate its adaptability in the more common and challenging domain of loosely-structured natural language reasoning.
To this end, we utilize the ProcessBench benchmark~\cite{Zheng2024ProcessBenchIP}, a dataset designed to evaluate models on a variety of reasoning errors.
Each test instance consists of a question, a step-by-step solution generated by a LLM, and a human-annotated label indicating the index of the first incorrect step. 
If the solution is entirely correct, the label is assigned as $-1$.
We use the full test sets from three ProcessBench subsets: \emph{GSM8K} (elementary-level math problems), \emph{MATH} (more advanced math problems), and \emph{OlympiadBench} (Olympiad-level math problems), comprising 400, 1000, and 1000 samples, respectively.
These subsets collectively span a broad range of reasoning difficulty, offering representative coverage of typical mathematical tasks.

To comprehensively evaluate GoV's performance, we compare it against two distinct verification paradigms:
\textbf{Holistic Verification:} This represents the standard practice, where the entire reasoning process is provided to an LLM verifier in a single pass. We use the official results from the ProcessBench leaderboard as our baseline.
\textbf{Decomposition with Premise Extraction (PARC):} As a state-of-the-art decomposition-based method, PARC first attempts to identify the minimal premises for each reasoning step before verification. This allows for a direct comparison of GoV's pragmatic approach against strategies that pursue premise minimality.

To address the challenges of loosely-structured reasoning, where atomic decomposition is brittle and impractical, we configured GoV for its high-robustness mode. This strategy, guided by our design space, entails:
\begin{itemize}
    \item \textbf{Block Granularity:} We treat each natural language paragraph as a node block ($B_j$). This choice leverages the text's inherent semantic structure, aligning our verification units with the author's logical stages.
    \item \textbf{Inclusive Context:} During the verification of a given block, we provide an inclusive context ($Pred_{\text{prov}}$), which consists of the full text of all previously validated paragraphs. This pragmatic approach ensures all potentially relevant information is available, critically avoiding the risks associated with a fragile premise extraction step.
\end{itemize}
As we will demonstrate, this robust configuration is key to GoV's superior performance on natural language tasks.

A cornerstone of GoV's power is its practical flexibility, which allows verification strategies to be customized to the native structure of the reasoning process. 
Crucially, this showcases that GoV circumvents the need for a rigid, pre-defined atomic-level DAG. Instead, its adaptable node block system is engineered to leverage the inherent, macroscopic logical structure found in various reasoning formats. The superiority of this flexible design is confirmed by our experimental results, where GoV consistently outperforms more rigid approaches.

As shown in Table~\ref{tab:process_res}, GoV achieves state-of-the-art performance, consistently and significantly outperforming all comparative methods across the various models and datasets.

\subsubsection{Superiority over Holistic Verification:} Compared to the holistic baseline, GoV's success confirms the fundamental benefit of decomposition, reducing cognitive load and enabling more focused, accurate verification. The pronounced gains on complex datasets like \emph{OlympiadBench} (e.g., F1 score from 35.9\% to 62.2\% for \emph{Qwen2.5-32B-Instruct}) highlight this advantage.
    
\subsubsection{Superiority over PARC:} More importantly, GoV consistently outperforms PARC, a fellow decomposition-based framework. 
We attribute this significant advantage directly to GoV's adaptive strategy. 
While PARC relies on a fragile, LLM-based step to extract minimal premises, GoV's Block Granularity and Inclusive Context strategy proves far more robust in the noisy context of natural language. 
This result provides powerful empirical evidence for our central thesis: for loosely-structured tasks, a verification strategy that prioritizes decompositional robustness over contextual minimality is superior.


In summary, these results demonstrate that GoV's adaptive, pragmatic approach to decomposition enables a new state of the art in training-free reasoning verification. 

\section{Conclusion}
\label{sec:conclusion}

In this work, we addressed the critical challenge of verifying complex LLM reasoning, where existing methods often struggle with the trade-off between precision and robustness across diverse reasoning structures. We introduced the Graph of Verification (GoV), a novel, training-free framework that models reasoning as a Directed Acyclic Graph and leverages a flexible node block architecture for adaptable, multi-granular verification.

Our primary contribution is the formalization of decomposed verification within a two-dimensional design space of Verification Granularity and Contextual Scope. Building on this, we developed GoV, a practical framework that can navigate this space, tailoring its verification strategy—from fine-grained atomic checks for highly-structured tasks to robust paragraph-level blocks for loosely-structured natural language narratives. Our empirical evaluation on contrasting benchmarks demonstrated GoV's superiority over both holistic baselines and other decomposition-based methods. The results validate our central hypothesis: adaptability is key to achieving reliable verification, and GoV's pragmatic approach of prioritizing robustness over contextual minimality is crucial for narrative reasoning.

While GoV establishes a new standard for training-free verification, we recognize avenues for future work. The current segmentation of reasoning into nodes or blocks relies on the task's inherent structure; developing automated methods to parse unstructured reasoning into optimal GoV graphs is a compelling research direction. Furthermore, extending GoV from a pure verifier to an interactive "Graph of Correction" framework, where it not only identifies but also proposes and validates fixes for flawed reasoning steps, presents an exciting frontier.

Ultimately, Graph of Verification is a significant step toward transparent, scrutable, and reliable AI. By providing a structured lens to inspect LLM reasoning, we pave the way for trust in not only the final answers but the processes that produce them.

\section*{Acknowledgments}
This work was supported by the National Natural Science Foundation of China (Grant No. 62506210), Taishan Scholars Program No.TSQN202507241, Key R\&D Program of Shandong Province, China, No.2025KJHZ013, Shandong Provincial University Youth Innovation and Technology Support Program No.2022KJ291, Shandong Provincial Natural Science Foundation for Young Scholars Program No.ZR2025QC1627 and Qilu University of Technology (Shandong Academy of Sciences) Youth Outstanding Talent Program No. 2024QZJH02.

\bibliography{ref}


\appendix
\renewcommand{\thesection}{\Alph{section}}
\renewcommand{\thesubsection}{\thesection.\arabic{subsection}}
\setcounter{secnumdepth}{1}
\section{Implementation Details}
\label{sec:implementation}

\subsection{Algorithmic Description}

The algorithms for GoV is summarized in Algorithm~\ref{alg:gov}.

\begin{algorithm}[ht]
\caption{Graph of Verification (GoV)}
\label{alg:gov}
\begin{algorithmic}[1]
\REQUIRE Reasoning graph $\mathcal{G} = (V, E)$, LLM-based verification function $\text{Verify}$
\ENSURE Verification outcome $T(x)$, $x$ can be a node  or a node block

\STATE Perform a topological sort on $\mathcal{G}$ to obtain an ordering $\sigma$ and the corresponding verification sequence $\mathcal{C}_\text{verif}$ (or $\mathcal{B}$)
\FOR{each $x$ in $\mathcal{C}_\text{verif}$}
    \STATE Construct $Pred_\text{prov}(x)$ as a superset of $Pred(x)$ for node-level or $Pred_\text{ext}(x)$ for node block-level
    \IF{$Pred_\text{prov}(x) = \emptyset$}
        \STATE $T(x) \gets \text{Verify}(x)$ \hfill // Directly verify initial premise
        \IF{$T(x) = \text{False}$}
            \STATE \textbf{break} \hfill // Stop at first detected error
        \ENDIF
    \ELSE
        \STATE $T(x) \gets \big( \bigwedge_{y \in Pred_\text{prov}(x)} T(y) = \text{True} \big) $ $\land \text{Verify}\big(x \mid Pred_\text{prov}(x)\big)$
        \IF{$T(x) = \text{False}$}
            \STATE \textbf{break} \hfill // Stop at first detected error
        \ENDIF
    \ENDIF
\ENDFOR
\STATE \textbf{return} $\{T(x) \mid x \in \mathcal{C}_\text{verif}\}$ \hfill // Overall validity of the reasoning
\end{algorithmic}
\end{algorithm}

\subsection{Prompt Template}
\label{app:prompt}

The prompt used in the baseline method is constructed based on that of ProcessBench, while the prompt for GoV is a modified version derived from it. 
The baseline verifies the entire reasoning process as a whole, whereas GoV performs iterative verification in a node-block-wise manner following the graph structure. 
This design ensures that each step in GoV focuses solely on verifying the correctness of a localized part of the reasoning. 
In particular, for the Number Triangle Summation task, since each addition formula already explicitly contains its prerequisite information, the \texttt{\{calculation\}} in GoV's prompt is restricted to the current addition formula only.

\begin{tcolorbox}[title = {Baseline Prompt}, breakable]

The following is a math problem and a solution (split into paragraphs, enclosed with tags and
indexed from 0):
\vspace{2mm}

\textbf{[Math Problem]} \\
\colorbox{gray!20}{\texttt{\{problem\}}}

\vspace{2mm}

\textbf{[Solution]}

\textless{}paragraph\_0\textgreater{}

\colorbox{gray!20}{\texttt{\{paragraph 0 of solution\}}}

\textless{}/paragraph\_0\textgreater{}

...

\textless{}paragraph\_n-1\textgreater{}

\colorbox{gray!20}{\texttt{\{paragraph n-1 of solution\}}}

\textless{}/paragraph\_n-1\textgreater{}

\vspace{2mm}

Your task is to review and critique the solution paragraph by paragraph. Once you identify an error in a paragraph, return the index of the paragraph where the earliest error occurs. Otherwise, return the index of -1 (which typically denotes "not found").
\vspace{2mm}

Please put your final answer (i.e., the index) in \text{\textbackslash boxed\{\}}.
\end{tcolorbox}



\begin{tcolorbox}[title={GoV Prompt: Number Triangle Summation}, breakable]
\raggedright
Please verify whether the calculation \colorbox{gray!20}{\texttt{\{calculation\}}} is correct, and put your final conclusion (i.e., 'correct' or 'incorrect') in \texttt{\textbackslash boxed\{\}}.
\end{tcolorbox}

\begin{tcolorbox}[title={GoV Prompt: ProcessBench}, breakable, enhanced]
The following is a math problem, the verified preceding steps of its solution, and the current step to be evaluated.

\vspace{2mm}

\textbf{[Math Problem]} \\
\colorbox{gray!20}{\texttt{\{problem\}}}

\vspace{2mm}

\textbf{[Correct Preceding Steps]} \\
\colorbox{gray!20}{\texttt{\{preceding\_text if preceding\_text}} \colorbox{gray!20}{\texttt{else "No preceding steps."\}}}

\vspace{2mm}

\textbf{[Current Step]} \\
\colorbox{gray!20}{\texttt{\{current\_step\_text\}}}

\vspace{2mm}

Your task is to judge whether the [Current Step] is correct. Once you identify an error in the [Current Step], return ``incorrect''. Otherwise, return ``correct''.

\vspace{2mm}

Please think step by step, and put your final conclusion (i.e., correct or incorrect) in \text{\textbackslash boxed\{\}}.
\end{tcolorbox}

\subsection{Flexible Configurations of GoV}
\label{app:gov_examples}


A key feature of the GoV framework lies in its flexible granularity of verification units, which enables it to adapt to the complexity and structure of different reasoning problems. 

To illustrate this, we present three examples corresponding to different types of reasoning tasks: arithmetic calculation (Problem 1), fractional refund estimation (Problem 2), and position tracking through sequential state transitions (Problem 3). Each problem is first decomposed into atomic reasoning steps (nodes), with logical dependencies encoded as edges in a DAG. We then demonstrate how different topological orders of the DAG and the granularity of the verification unit division can be used to guide the verification process in a localized and structured manner.

\begin{reasoningbox}[label={q1}, title={Problem 1: Mario's Snowshoes}]

\textbf{Problem:}\\
Mario needs to buy snowshoes for his 6 sled dogs. Assuming his dogs each has four legs and each pair of snowshoes costs \$12.00, how much will it cost him to buy snowshoes for all of his dogs?

\vspace{1mm}
\textbf{Solution:}

Let's break down the problem step by step. 

First, each dog has 4 legs.

Second, Mario has 6 dogs, so in total, there are 6 x 4 = 24 legs.

Third, since each pair of snowshoes is for one leg, Mario needs 24 pairs of snowshoes.

Fourth, each pair of snowshoes costs \$12.00, so the total cost is 24 x \$12.00.

To calculate the total cost, we multiply 24 by 12: 24 x 12 = 288. 

So, Mario needs to spend $\boxed{\$288.00}$ to buy snowshoes for all his dogs.

\vspace{1mm}
\textbf{Reasoning Structure:}
\begin{itemize}[leftmargin=*,nosep]
  \item P1: Each dog has 4 legs
  \item P2: Mario has 6 dogs
  \item P3: Each pair of snowshoes costs \$12.00
  \item C1: Total legs = 6 $\times$ 4 = 24
  \item C2: Need 24 pairs
  \item C3: Cost = 24 $\times$ \$12 = \$288
\end{itemize}

\vspace{1mm}
\textbf{Dependencies:}
\begin{itemize}[leftmargin=*,nosep]
  \item P1 $\rightarrow$ C1
  \item P2 $\rightarrow$ C1
  \item C1 $\rightarrow$ C2
  \item P3 $\rightarrow$ C3
  \item C2 $\rightarrow$ C3
\end{itemize}

\vspace{1mm}
\textbf{Topological Orders:}
\begin{itemize}[leftmargin=*,nosep]
  \item P1, P2, P3, C1, C2, C3
  \item P1, P2, C1, P3, C2, C3 \textbf{(Selected)}
  \item P1, P2, C1, C2, P3, C3
  \item P1, P3, P2, C1, C2, C3
  \item P2, P1, P3, C1, C2, C3
  \item P2, P1, C1, P3, C2, C3
  \item P2, P1, C1, C2, P3, C3
  \item P2, P3, P1, C1, C2, C3
  \item P3, P1, P2, C1, C2, C3
  \item P3, P2, P1, C1, C2, C3
\end{itemize}

\vspace{1mm}
\textbf{Verification Steps (Node-Level):}
\begin{itemize}[leftmargin=*,nosep]
  \item Verify P1, P2, P3 from problem
  \item Verify C1 from P1 and P2
  \item Verify C2 from C1
  \item Verify C3 from C2 and P3
  \item Verify if C3 fully answers the problem
\end{itemize}

\vspace{1mm}
\textbf{Node Block:}
\begin{itemize}[leftmargin=*,nosep]
  \item B1 = [P1, P2]
  \item B2 = [C1]
  \item B3 = [P3]
  \item B4 = [C2]
  \item B5 = [C3]
\end{itemize}

\vspace{1mm}
\textbf{Verification Steps (Node Block-Level):}
\begin{itemize}[leftmargin=*,nosep]
  \item Verify B1 based on Problem
  \item Verify B2 based on B1
  \item Verify B3 based on Problem
  \item Verify B4 based on B2
  \item Verify B5 based on B3 and B4
  \item Verify whether the Problem is fully answered based on Problem and B5
\end{itemize}
\end{reasoningbox}

\begin{reasoningbox}[label={q2}, title={Problem 2: Milk Refund}]

\textbf{Problem:}\\
A company sold 4000 gallons of milk in jars to Mr. Marcellus' store at the cost of \$3.5 per gallon. However, Mr. Marcellus later realized 2/5 of the amount of milk he purchased had passed the expiry date and could not be sold. He returned the sour milk to the company and ordered a refund. Calculate how much he got in refunds.

\vspace{1mm}
\textbf{Solution:}

First, let's calculate the total cost of the milk that Mr. Marcellus bought: Cost per gallon = \$3.5, Total gallons purchased = 4000. Total cost = Cost per gallon * Total gallons purchased = \$3.5 * 4000 = \$14,000.

Next, we need to find out how much milk was returned due to expiration: Fraction of expired milk = 2/5. Amount of expired milk = Fraction of expired milk * Total gallons purchased = 2/5 * 4000 = 800 gallons.

Now, we need to calculate the cost of the expired milk that needs to be refunded: Cost of expired milk = Cost per gallon * Amount of expired milk = \$3.5 * 800 = \$2,800.

Therefore, Mr. Marcellus got a refund of \$2,800 for the unsellable milk. The answer is $\boxed{2800}$.

\vspace{1mm}
\textbf{Reasoning Structure:}
\begin{itemize}[leftmargin=*,nosep]
    \item P1: \$3.5 per gallon
    \item P2: 4000 gallons purchased
    \item P3: 2/5 expired
    \item C1: Total cost = \$14{,}000
    \item C2: Expired milk = 800 gallons
    \item C3: Refund = \$2800
\end{itemize}

\vspace{1mm}
\textbf{Dependencies:}
\begin{itemize}[leftmargin=*,nosep]
    \item P1 $\rightarrow$ C1
    \item P2 $\rightarrow$ C1
    \item P2 $\rightarrow$ C2
    \item P3 $\rightarrow$ C2
    \item P1 $\rightarrow$ C3
    \item C2 $\rightarrow$ C3
\end{itemize}

\vspace{1mm}
\textbf{Topological Orders:}
\begin{itemize}[leftmargin=*,nosep]
    \item P1, P2, P3, C1, C2, C3 \textbf{(Selected)}
    \item P1, P2, P3, C2, C1, C3
    \item P1, P3, P2, C1, C2, C3
    \item P1, P3, P2, C2, C1, C3
    \item P2, P1, P3, C1, C2, C3
    \item P2, P1, P3, C2, C1, C3
    \item P2, P3, P1, C1, C2, C3
    \item P2, P3, P1, C2, C1, C3
    \item P3, P1, P2, C1, C2, C3
    \item P3, P1, P2, C2, C1, C3
    \item P3, P2, P1, C1, C2, C3
    \item P3, P2, P1, C2, C1, C3
\end{itemize}

\vspace{1mm}
\textbf{Verification Steps:}
\begin{itemize}[leftmargin=*,nosep]
    \item Verify P1 based on Problem
    \item Verify P2 based on Problem
    \item Verify P3 based on Problem
    \item Verify C1 based on P1 and P2
    \item Verify C2 based on P2 and P3
    \item Verify C3 based on P1 and C2
    \item Verify whether the Problem is fully answered based on Problem and C3
\end{itemize}
\end{reasoningbox}

\begin{reasoningbox}[label={q3}, title={Problem 3: Finley's Race Position}]

\textbf{Problem:}\\
Finley took part in a 100-meter race. She started off in first, but then fell back 5 spots. She then moved ahead 2 spots, before falling behind 3. Lastly, she jumped ahead 1 spot to finish the race. What place did she finish in?

\vspace{1mm}
\textbf{Solution:}

To find out what place Finley finished in, let's break down her movements step by step: 

First, Finley started off in first place. - Current position: 1st

Second, she fell back 5 spots. - New position: 1st - 5 = 6th

Third, she moved ahead 2 spots. - New position: 6th + 2 = 8th

Fourth, she fell behind 3 spots. - New position: 8th - 3 = 11th

Lastly, she jumped ahead 1 spot to finish the race. - New position: 11th + 1 = 12th

So, Finley finished the race in 12th place. \boxed{12}

\vspace{1mm}
\textbf{Reasoning Structure:}
\begin{itemize}[leftmargin=*,nosep]
    \item P1: Start at 1st
    \item P2: Fall back 5
    \item P3: Move ahead 2
    \item P4: Fall behind 3
    \item P5: Jump ahead 1
    \item C1: 6th
    \item C2: 8th
    \item C3: 11th
    \item C4: 12th
\end{itemize}

\vspace{1mm}
\textbf{Dependencies:}
\begin{itemize}[leftmargin=*,nosep]
    \item P1 $\rightarrow$ C1
    \item P2 $\rightarrow$ C1
    \item C1 $\rightarrow$ C2
    \item P3 $\rightarrow$ C2
    \item C2 $\rightarrow$ C3
    \item P4 $\rightarrow$ C3
    \item C3 $\rightarrow$ C4
    \item P5 $\rightarrow$ C4
\end{itemize}

\vspace{1mm}
\textbf{Topological Orders:}
\begin{itemize}[leftmargin=*,nosep]
    \item P1, P2, C1, P3, C2, P4, C3, P5, C4 \textbf{(Selected)} and 119 others
\end{itemize}

\vspace{1mm}
\textbf{Verification Steps:}
\begin{itemize}[leftmargin=*,nosep]
    \item Verify P1 based on Problem
    \item Verify P2 based on Problem
    \item Verify C1 based on P1 and P2
    \item Verify P3 based on Problem
    \item Verify P4 based on Problem
    \item Verify C3 based on P4 and C2
    \item Verify P5 based on Problem
    \item Verify C4 based on P5 and C3
    \item Verify whether the Problem is fully answered based on Problem and C4
\end{itemize}
\end{reasoningbox}


\section{Dataset Details}
\label{sec:dataset}
This section provides data examples for the Number Triangle Summation and ProcessBench datasets.

\subsection{Number Triangle Summation}
\label{app:nts_detail}
An error in any intermediate calculation typically propagates, leading to an incorrect final result. The `label` for this dataset indicates the first step (1-indexed) containing a calculation error.
\begin{tcolorbox}[breakable, title={Example: Number Triangle Summation}]
\small
\textbf{Problem:} \\
Given 5 numbers at layer 0: 8129, 1967, 7829, 4593, 4076. Calculate the sum of adjacent numbers to form the next layer, repeating this process until only one number remains at layer 4. What is the final number?
\vspace{1mm}

\textbf{Solution:} \\
Step 1 (Calculating Layer 1 from Layer 0):\\
  8129 + 1967 = 10096\\
  1967 + 7829 = 9796\\
  7829 + 4593 = 12422\\
  4593 + 4076 = 8669
\vspace{1mm}

Step 2 (Calculating Layer 2 from Layer 1):\\
  10096 + 9796 = 19892\\
  9796 + 12422 = 22218\\
  12422 + 8669 = 21091
\vspace{1mm}

Step 3 (Calculating Layer 3 from Layer 2):\\
  19892 + 22218 = 42110\\
  \textcolor{red}{22218 + 21091 = 43209} \textit{\quad\% Correct sum is 43309}
\vspace{1mm}

Step 4 (Calculating Layer 4 from Layer 3):\\
  42110 + 43209 = 85319
\vspace{1mm}

Final Answer: The number at Layer 4 is 85319.
\vspace{2mm}

\textbf{Final Answer Correct:} \texttt{FALSE} \\
\textbf{Label:} \texttt{3}
\end{tcolorbox}

\subsection{ProcessBench}
For ProcessBench, the solution is presented in paragraphs, and the `label` indicates the 0-indexed paragraph containing the first error. A label of -1 signifies a correct solution.
\begin{tcolorbox}[breakable, title={Example: ProcessBench (GSM8K subset, Erroneous)}]
\small
\textbf{Problem:} \\
Mario needs to buy snowshoes for his 6 sled dogs. Assuming his dogs each has four legs and each pair of snowshoes costs \$12.00, how much will it cost him to buy snowshoes for all of his dogs?
\vspace{1mm}

\textbf{Solution:} \\
\textit{(Paragraph 0)} Let's break down the problem step by step. First, each dog has 4 legs.
\vspace{1mm}

\textit{(Paragraph 1)} Second, Mario has 6 dogs, so in total, there are 6 x 4 = 24 legs.
\vspace{1mm}

\textcolor{red}{\textit{(Paragraph 2)} Third, since each pair of snowshoes is for one leg, Mario needs 24 pairs of snowshoes.} \textit{\quad\% Error: A pair is for two legs.}
\vspace{1mm}

\textit{(Paragraph 3)} Fourth, each pair of snowshoes costs \$12.00, so the total cost is 24 x \$12.00.
\vspace{1mm}

\textit{(Paragraph 4)} To calculate the total cost, we multiply 24 by 12: 24 x 12 = 288. So, Mario needs to spend \$\boxed{288.00}\$ to buy snowshoes for all his dogs.
\vspace{2mm}

\textbf{Final Answer Correct:} \texttt{FALSE} \\
\textbf{Label:} \texttt{2}
\end{tcolorbox}


\section{Cost and Efficiency Analysis}
\label{app:cost}

The inherently structured and iterative verification methodology of GoV, characterized by independent LLM calls for each node or node block, naturally incurs greater computational resource consumption compared to holistic, single-pass verification techniques such as the CoT baseline. This section provides a transparent analysis of token consumption and corresponding monetary costs for the \emph{Qwen2.5-32B-Instruct} model applied to the Number Triangle Summation and GSM8K (ProcessBench) datasets. The pricing is approximately \$0.0003 per 1,000 input tokens and \$0.0009 per 1,000 output tokens.

Table~\ref{tab:cost_number_triangle_appendix} outlines the cost comparison for the Number Triangle Summation task. The significantly higher output token volume for GoV stems from the LLM's need to generate detailed, step-by-step verification reasoning for the numerous atomic arithmetic operations within the number triangle.

\begin{table}[h]
\centering
\resizebox{\columnwidth}{!}{%
\begin{tabular}{lrrr}
\specialrule{1.5pt}{0pt}{0pt}
& \textbf{Input Tokens} & \textbf{Output Tokens} & \textbf{Total Cost (\$)} \\
\midrule
CoT (Baseline) & 237,654 & 76,622 & 0.140 \\
GoV (Ours) & 264,687 & 1,388,141 & 1.329 \\
\specialrule{1.5pt}{0pt}{0pt}
\end{tabular}%
}
\caption{Cost analysis for the entire Number Triangle Summation dataset (500 instances).}
\label{tab:cost_number_triangle_appendix}
\end{table}

For the GSM8K (ProcessBench) dataset, the cost implications are detailed in Table~\ref{tab:cost_gsm8k_processbench_appendix}. The total cost of verifying these 400 instances using GoV is approximately \$0.362, compared to \$0.101 for the baseline. In this context, GoV's paragraph-level node block verification strategy results in a higher input token count due to the need to re-provide previously verified paragraphs as context for subsequent blocks. Concurrently, the iterative verification of each distinct block also leads to increased output tokens.

\begin{table}[h]
\centering
\resizebox{\columnwidth}{!}{%
\begin{tabular}{lrrr}
\specialrule{1.5pt}{0pt}{0pt}
& \textbf{Input Tokens} & \textbf{Output Tokens} & \textbf{Total Cost (\$)} \\
\midrule
CoT (Baseline) & 210,860 & 42,157 & 0.101 \\
GoV (Ours) & 578,592 & 208,820 & 0.362 \\
\specialrule{1.5pt}{0pt}{0pt}
\end{tabular}%
}
\caption{Cost analysis for the entire GSM8K (ProcessBench) dataset (400 instances).}
\label{tab:cost_gsm8k_processbench_appendix}
\end{table}

While the experiments above show GoV's cost to be 3 to 10 times higher than the CoT baseline, GoV's flexibility allows for the aggregation of nodes into node blocks, which can reduce verification costs and improve efficiency. Consequently, GoV can be configured according to budgetary constraints; in an extreme case where all nodes are aggregated into a single node block, GoV effectively degenerates into the CoT baseline.

There is a growing recognition that achieving superior AI performance, particularly in complex cognitive tasks, often necessitates more intensive and structured computational approaches. As LLM technology continues to advance, potentially bringing higher operational efficiency and lower costs, the strategic advantages of adopting rigorous verification frameworks like GoV are expected to become more compelling and integral to the development of reliable and trustworthy AI systems.

\section{More Results}

\label{app:more_results}
For completeness, we also include the performance of Process Reward Models (PRMs) as reported in their original paper. PRMs are fine-tuned verifiers trained on human-annotated reasoning steps, representing a fundamentally different paradigm from our proposed method. While GoV is entirely training-free and model-agnostic, PRMs rely on supervised learning with task-specific labels and dedicated model optimization. We include PRM as a strong reference point to contextualize the performance of GoV, despite the methodological differences.

Table~\ref{tab:prm_perform} presents the results of PRM across several model sizes, alongside the corresponding results of GoV using models of the same size. As shown, GoV consistently achieves significantly better performance despite being entirely training-free. 
The advantage is particularly pronounced on the more challenging \emph{OlympiadBench}. 
In contrast, PRM relies on supervised fine-tuning with human-annotated reasoning steps. This highlights the strong practical advantage of GoV in terms of both effectiveness and adaptability, without the need for additional training or labeled data.
\newpage
\onecolumn

\begin{table*}[t]
\centering
\resizebox{\linewidth}{!}{
\begin{tabular}{ccccccccccc}
\specialrule{1.5pt}{0pt}{0pt}
\multirow{2}{*}{\textbf{Model}} & \multirow{2}{*}{\textbf{Method}} & \multicolumn{3}{c}{\textbf{GSM8K}} & \multicolumn{3}{c}{\textbf{MATH}} & \multicolumn{3}{c}{\textbf{OlympiadBench}} \\ \cmidrule(lr){3-5} \cmidrule(lr){6-8} \cmidrule(lr){9-11}
 &  & Correct Acc & Wrong Acc & \textbf{F1} & Correct Acc & Wrong Acc & \textbf{F1} & Correct Acc & Wrong Acc & \textbf{F1} \\ \specialrule{1.15pt}{0pt}{0pt}
\multirow{4}{*}{\begin{tabular}[c]{@{}c@{}}PRM\end{tabular}} 
 & Math-Shepherd-PRM-7B & 32.4 & 91.7 & 47.9 & 18.0 & 82.0 & 29.5 & 15.0 & 71.1 & 24.8 \\
 & RLHFlow-PRM-Mistral-8B & 33.8 & 99.0 & 50.4 & 21.7 & 72.2 & 33.4 & 8.2 & 43.1 & 13.8 \\
 & RLHFlow-PRM-Deepseek-8B  & 24.2 & 98.4 & 38.8 & 21.4 & 80.0 & 33.8 & 10.1 & 51.0 & 16.9\\
 & Skywork-PRM-7B  & 61.8 & 82.9 & \textbf{70.8} & 43.8 & 69.2 & 53.6 & 17.9 & 31.9 & 22.9\\\hline
 
 GoV & Qwen2.5-7B-Instruct & 89.1 & 43.0 & 58.0 & 80.7 & 41.7 & \textbf{55.0} & 74.9 & 30.1 & \textbf{42.9} \\ \specialrule{1.5pt}{0pt}{0pt}

\end{tabular}
}
\caption{Performance comparison on ProcessBench.}
\label{tab:prm_perform}
\end{table*}

\section{Case Study}
\label{app:casestudy}

To clearly illustrate the strengths of the GoV approach, Table~\ref{tab:snowshoes_case_study} presents a representative case from a \emph{GSM8K} subset. 
In this example, the reasoning process contains a subtle logical flaw. 
The baseline method, which attempts to handle the entire reasoning chain in one pass, fails to detect the error and instead incorrectly classifies the output as correct. 
In contrast, GoV identifies the flawed logic within the specific paragraph responsible for the error. 
By focusing on this segment while building upon the previously validated context, GoV enables the model to isolate and flag the mistake effectively. 
This example highlights how GoV's step-by-step verification mechanism empowers LLMs to conduct more fine-grained and reliable reasoning, capturing subtle defects that are often overlooked in holistic evaluations.

\begin{longtable}{@{} >{\raggedright\arraybackslash}p{2.5cm} p{4cm} p{8cm} @{}}
\specialrule{1.5pt}{0pt}{0pt}
\textbf{Category} & \textbf{Component} & \textbf{Details} \\
\specialrule{0.6pt}{0pt}{0pt}
\endfirsthead

\multicolumn{3}{c}%
{{\tablename\ \thetable{} -- Continued from previous page}} \\
\toprule
\textbf{Category} & \textbf{Component} & \textbf{Details} \\
\midrule
\endhead

\midrule \multicolumn{3}{r}{{Continued on next page}} \\
\endfoot

\bottomrule
\caption{Analysis of LLM Solution for Mario's Snowshoes Problem} \label{tab:snowshoes_case_study} \\
\endlastfoot

\multicolumn{2}{@{}l}{\multirow{4}{*}{\textbf{Problem}}}  & Mario needs to buy snowshoes for his 6 sled dogs. Assuming his dogs each has four legs and each pair of snowshoes costs \$12.00, how much will it cost him to buy snowshoes for all of his dogs? \\
\midrule

\multirow{11}{*}{\textbf{LLM's Solution}}
& Block $B_1$ & [Let's break down the problem step by step. First, each dog has 4 legs.] \\
\cmidrule(l){2-3}
& Block $B_2$ & [Second, Mario has 6 dogs, so in total, there are 6 x 4 = 24 legs.] \\
\cmidrule(l){2-3}
& Block $B_3$ & [Third, since each pair of snowshoes is for one leg, Mario needs 24 pairs of snowshoes.] \\
\cmidrule(l){2-3}
& Block $B_4$ & [Fourth, each pair of snowshoes costs \$12.00, so the total cost is 24 x \$12.00.] \\
\cmidrule(l){2-3}
& Block $B_5$ & [To calculate the total cost, we multiply 24 by 12: 24 x 12 = 288. So, Mario needs to spend $\boxed{\$288.00}$ to buy snowshoes for all his dogs.] \\
\midrule

\multirow{20}{=}{\textbf{Baseline (COT) Verification}}
& \multirow{19}{=}{Reason} & [Reading through the provided solution step by step: \par \textless paragraph\_0\textgreater accurately states that each dog has 4 legs, which is correct. \par \textless paragraph\_1\textgreater correctly calculates that for 6 dogs, there are a total of 6 x 4 = 24 legs, which is accurate. \par \textless paragraph\_2\textgreater correctly deduces that since each pair of snowshoes covers one leg, Mario needs 24 pairs of snowshoes, which is accurate based on the previous steps. \par  \colorbox{red!20}{\textless paragraph\_3\textgreater correctly notes that each pair of snow-} \colorbox{red!20}{shoes costs \$12.00, and the total cost can be found by} \colorbox{red!20}{multiplying the number of pairs by the cost, which is} \colorbox{red!20}{24 x \$12.00.} \textcolor{red}{\ding{55}} \par \textless paragraph\_4\textgreater correctly calculates the total cost as 24 x \$12.00 = \$288.00. \par After reviewing the solution paragraph by paragraph, there are no errors found in the logic or calculations. Therefore, the index of the paragraph where the earliest error occurs is \boxed{-1}.] \\
\cmidrule(l){2-3}
& Judgment & \colorbox{red!20}{[Correct]} \textcolor{red}{\ding{55}}\\
\midrule

\multirow{43}{=}{\textbf{GoV Verification Process}}
& \multirow{10}{=}{Step 1: Verify Block $B_1$} & \textbf{Premises:} (None or initial problem statement) \par \textbf{Reason:} [The current step is stating a fact about the number of legs each dog has, which is correct according to the problem statement. The problem mentions that each dog has four legs, and this is the first step in understanding the total number of snowshoes needed. \par Therefore, the evaluation of the [Current Step] is \boxed{correct}. ] \par \textbf{LLM Judgment on $B_1$:} Correct \\
\cmidrule(l){2-3}
& \multirow{13}{=}{Step 2: Verify Block $B_2$} & \textbf{Premises:} Content of $B_1$ \par \textbf{Reason:} [To evaluate the correctness of the [Current Step], let's go through the logic step by step. \par 1. Each dog has 4 legs. \par 2. Mario has 6 dogs. \par 3. Therefore, the total number of legs is 6 dogs * 4 legs/dog = 24 legs. \par The [Current Step] states that Mario has 6 dogs, so in total, there are 6 x 4 = 24 legs. This calculation is correct based on the given information. \par Therefore, the evaluation of the [Current Step] is \boxed{correct}. ] \par \textbf{LLM Judgment on $B_2$:} Correct \\
\cmidrule(l){2-3}
& \multirow{19}{=}{Step 3: Verify Block $B_3$} & \textbf{Premises:} Content of $B_1$, $B_2$ \par \textbf{Reason:} [Let's analyze the [Current Step] step by step. \par 1. Each dog has 4 legs. \par 2. Mario has 6 dogs, so there are 6 x 4 = 24 legs in total. \par 3. The [Current Step] states that Mario needs 24 pairs of snowshoes, assuming each pair of snowshoes is for one leg. \par However, the term "pair" typically implies two items. In the context of snowshoes, a pair would be for two legs (one for each foot). Therefore, Mario would need 12 pairs of snowshoes to cover all 24 legs (since each pair covers two legs). \par Thus, the [Current Step] is incorrect because it assumes each pair of snowshoes is for one leg, which is not the standard usage of the term "pair." \par \boxed{incorrect}] \par \textbf{LLM Judgment on $B_3$:} INCORRECT .]\\
\cmidrule(l){2-3}
& GoV Conclusion & \colorbox{green!20}{==\textgreater{} GoV reports error at Block $B_3$ and stops.} \textcolor{green}{\ding{51}}\\

\end{longtable}

\end{document}